\documentclass[sigconf]{acmart}
\usepackage{hyperref}

\AtBeginDocument{%
  \providecommand\BibTeX{{%
    \normalfont B\kern-0.5em{\scshape i\kern-0.25em b}\kern-0.8em\TeX}}}

\copyrightyear{2020}
\acmYear{2020}
\setcopyright{acmcopyright}
\acmConference[MM '20] {28th ACM International Conference on Multimedia}{October 12--16, 2020}{Seattle, WA, USA.}
\acmBooktitle{28th ACM International Conference on Multimedia (MM '20), October 12--16, 2020, Seattle, WA, USA.}
\acmPrice{15.00}
\acmDOI{10.1145/3394171.3414040}
\acmISBN{978-1-4503-7988-5/20/10} 
\begin{document}

\title{Alleviating Human-level Shift : A Robust Domain Adaptation Method for Multi-person Pose Estimation}

\author{Xixia Xu}
\affiliation{
  \institution{Beijing Key Laboratory of Traffic Data Analysis and Mining, Beijing Jiaotong University, Beijing, 100044}
}
\email{19112036@bjtu.edu.cn}

\author{Qi Zou}
\authornote{Corresponding author.}
\affiliation{
  \institution{Beijing Key Laboratory of Traffic Data Analysis and Mining, Beijing Jiaotong University, Beijing, 100044}
}
\email{qzou@bjtu.edu.cn}

\author{Xue Lin}
\affiliation{
  \institution{Beijing Key Laboratory of Traffic Data Analysis and Mining, Beijing Jiaotong University, Beijing, 100044}
}
\email{18112028@bjtu.edu.cn}

\renewcommand{\shortauthors}{Xu and Zou, et al.}

\begin{abstract}
  Human pose estimation has been widely studied with much focus on supervised learning requiring sufficient annotations. However, in real applications, a pretrained pose estimation model usually need be adapted to a novel domain with no labels or sparse labels. Such domain adaptation for 2D pose estimation hasn’t been explored. The main reason is that a pose, by nature, has typical topological structure and needs fine-grained features in local keypoints. While existing adaptation methods do not consider topological structure of object-of-interest and they align the whole images coarsely. Therefore, we propose a novel domain adaptation method for multi-person pose estimation to conduct the human-level topological structure alignment and fine-grained feature alignment. Our method consists of three modules: Cross-Attentive Feature Alignment (CAFA), Intra-domain Structure Adaptation (ISA) and Inter-domain Human-Topology Alignment (IHTA) module. The CAFA adopts a bidirectional spatial attention module (BSAM) that focuses on fine-grained local feature correlation between two humans to adaptively aggregate consistent features for adaptation. We adopt ISA only in semi-supervised domain adaptation (SSDA) to exploit the corresponding keypoint semantic relationship for reducing the intra-domain bias. Most importantly, we propose an IHTA to learn more domain-invariant human topological representation for reducing the inter-domain discrepancy. We model the human topological structure via the graph convolution network (GCN), by passing messages on which, high-order relations can be considered. This structure preserving alignment based on GCN is beneficial to the occluded or extreme pose inference. Extensive experiments are conducted on two popular benchmarks and results demonstrate the competency of our method compared with existing supervised approaches. Code is avaliable on \url{https://github.com/Sophie-Xu/Pose_DomainAdaption}.
\end{abstract}

\ccsdesc[500]{Computing methodologies}
\ccsdesc[500]{Computing methodologies~Activity recognition and understanding}
\ccsdesc[300]{Computing methodologies~Transfer learning}

\keywords{Multi-person Pose Estimation; Domain Adaptation; Human-level Knowledge Alignment}

\maketitle
\begin{figure}[htp]
\centering
\includegraphics[width=3.2in,height = 1.8in]{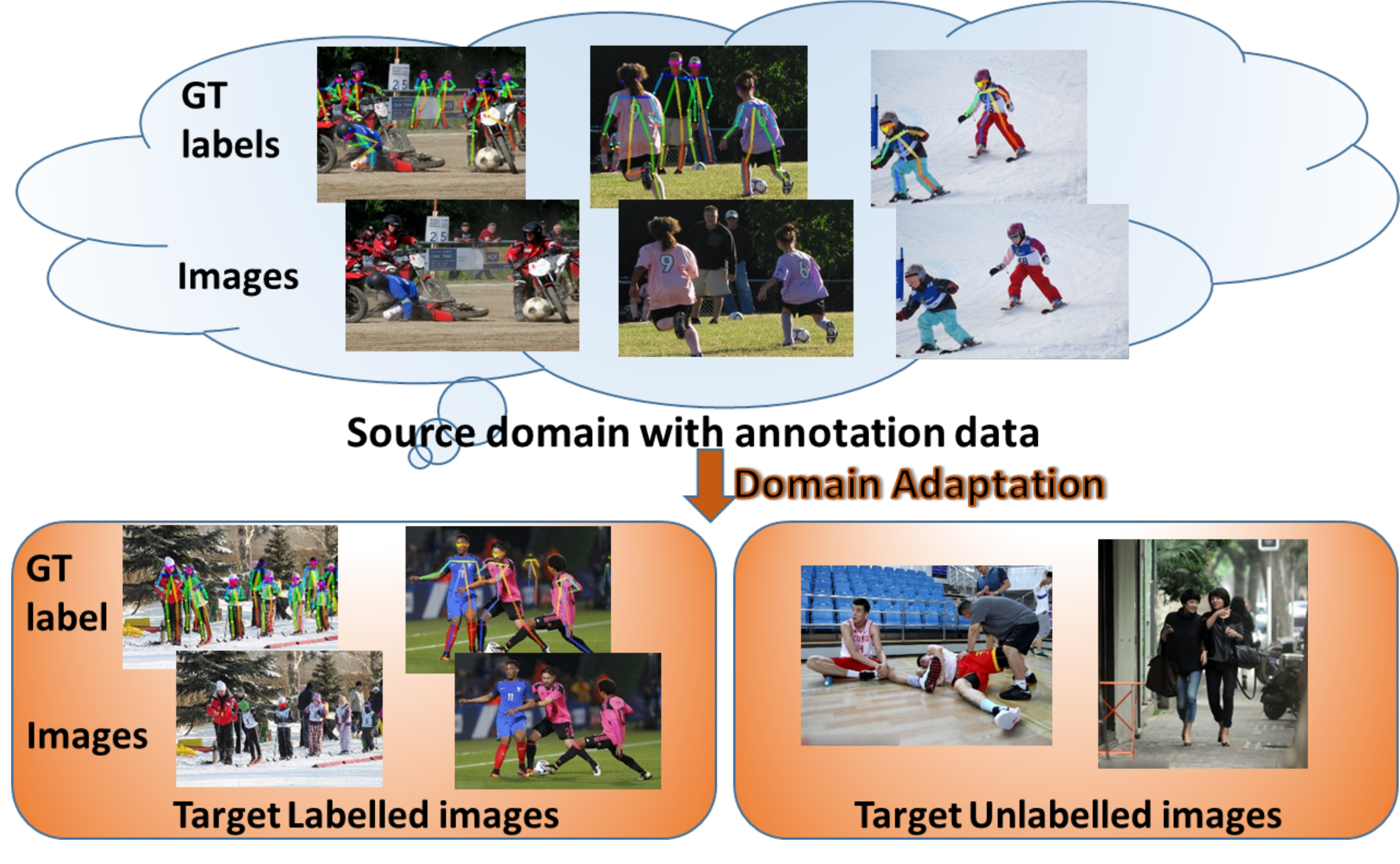}
\caption{The examples with visual domain gap between the source and target domain.}
\label{fig:1 st}
\end{figure}
\section{Introduction}
Multi-person pose estimation devotes to locate body parts for multiple persons in 2D image, such as keypoints on the arms, torsos, and the face\cite{2019Pose}. It's fundamental to deal with other high-level tasks, such as human action recognition\cite{Choutas2018PoTion} and human-computer interaction\cite{pantic2005affective}. Recently, due to the progress of convolution neural network (CNN)\cite{he2016deep}, most existing methods\cite{chen2018cascaded,li2018crowdpose:,2019Pose,xiao2018simple} have achieved remarkable advances in multi-person pose estimation. However, existing supervised methods cannot generalize well to a novel domain without labels or with sparse labels, especially when the new domain has a different distribution. A natural remedy is the unsupervised domain adaptation (UDA). UDA has been widely applied in computer vision field, such as image classification\cite{long2018conditional}, object detection\cite{saito2019strong-weak}, and semantic segmentation
\cite{hoffman2017cycada:}. In these cases, a model trained on a source domain with full annotations is adapted for an unlabeled target domain via minimizing the distribution discrepancy of the features\cite{long2015learning} or discriminating the output through adversarial learning\cite{pan2020adversarial,ganin2016domain-adversarial,tsai2019domain}. However, UDA for 2D pose estimation has never been explored. And applying the above domain adaptation methods into cross-domain pose estimation cannot guarantee the satisfying performance.

There are three types of adaptation challenges needed to be mitigated for cross-domain pose estimation: $\textbf{1)}$ Pose estimation needs fine-grained local features. However, how to adapt these consistent fine-grained human body features across domains remains unexplored. As illustrated in Fig~\ref{fig:1 st}, humans in source and target images share much similar semantic representations such as postures, scales and actions, although their surrounding environments are quite different. It can be seen that feature adaptation at the image-level will meet difficulties under such cases. Existing domain adaptation methods widely used in image classification and semantic segmentation\cite{hoffman2017cycada:,tsai2018learning} typically consider the image as a whole for alignment, while ignore local regions of the object-of-interest. Focusing on such local regions is important for pose estimation. Although some cross-domain object detection methods\cite{zhu2019adapting} focus on local objects, they aim at bridging the domain gap at a coarse granularity but not at the fine-grained keypoint level. Moreover, they often consider the domain features separately and neglect the local feature dependency across domains.   
$\textbf{2)}$ Human topological structure is pivotal for adaptation performance especially in extreme poses or occlusions, while it’s unrevealed yet. Existing domain adaptation methods for structural output\cite{tsai2019domain,tsai2018learning} adapt the output distribution and align the global layout across domains for semantic segmentation, which is much different from the human structure alignment in pose estimation. 
$\textbf{3)}$ In semi-supervised setting, the gap between labeled and unlabeled data in the target domain also exists, as shown in Fig~\ref{fig:1 st}. Few works\cite{kim2020cross-domain,luo2017label} consider the intra-domain gap via adapting feature representations simply, let alone exploring the intra-domain structural keypoint relationship.

Aiming at these issues, we propose a novel domain adaptation framework as shown in Fig~\ref{fig:2network}, which consists of three adaptation parts: $(1)$ $\textbf{C}$ross-$\textbf{A}$ttentive $\textbf{F}$eature $\textbf{A}$lignment $\textbf{(CAFA})$. To explore the similar fine-grained human body features and capture domain-invariant semantics across domains, we innovatively adopt a bidirectional spatial attention module (BSAM) to capture local feature similarity across humans. The local features of source human parts can be encoded in the target domain, and vice versa. It allows us to adaptively capture the consistent fine-grained human body features for adaptation.
$(2)$ $\textbf{I}$ntra-domain $\textbf{S}$tructure $\textbf{A}$daptation $\textbf{(ISA)}$. In SSDA setting, we exploit the annotations available in the target domain to learn the structural keypoint information, and adapt the reliable keypoint knowledge to the unlabeled data. Specifically, we align the one-to-one specific keypoint heatmap representation between the labeled and unlabeled data in target domain to augment the latter.
$(3)$ $\textbf{I}$nter-domain $\textbf{H}$uman-$\textbf{T}$opology $\textbf{A}$lignment $\textbf{(IHTA)}$. We adopt the recent SemGCN\cite{zhao2019semantic} to capture flexible human-topology representations. Besides, a sample selection mechanism is designed to determine which pairs should be aligned to avoid the hard alignment between arbitrary poses. Upon this, we align the human-topology representations to preserve structure-invariant knowledge across domains. The predicted errors in the target domain can be further repaired by the learned structure knowledge (e.g., the structural reasoning helps to infer the occluded or invisible poses).

Up to our knowledge, this is the first attempt that domain adaptation is explored under the multi-person pose estimation task. Our main contributions are as follows.

$\bullet$ A novel CAFA module achieves the fine-grained human body feature alignment and adapts abundant domain-invariant features for accurate pose estimation. We are also the first to investigate the transferability of fine-grained features via exploring the bidirectional spatial feature dependency across domains.

$\bullet$ In SSDA setting, a novel ISA adapts the local keypoint structural knowledge of the labeled to the unlabeled data in the target domain. It encourages the former to augment the keypoint representation of the latter to alleviate intra-domain confusions.

$\bullet$ The IHTA mechanism conducts human topological structure alignment softly to explicitly preserve high-order structure-invariant knowledge across domains. We additionally adopt a semantic graph-based formulation for modeling the human-topology information. 

$\bullet$ Comprehensive experiments demonstrate the competency of the proposed method compared with the existing supervised approaches on two benchmark settings, i.e., "MPII to MS-COCO" and "MS-COCO to MPII".
\section{Related Work}

\textbf{Supervised Multi-Person Pose Estimation.}
Recently, multi-pers
on pose estimation has aroused a great interest due to the real-life demand. Nowadays, researchers have made painstaking efforts \cite{2019Pose,ke2018multi-scale,sun2019deep,nie2019single-stage} to accelerate its progress. For examples, CASNet\cite{2019Pose} improves the feature representation via adopting the spatial and channel-wise attention. HRNet\cite{sun2019deep} builds a new strong baseline via elaborated network design. However, they are all trained on adequate labeled images. Very few works explore the weakly/semi-supervised study in this field. The PoseWarper\cite{bertasius2019learning} leverages the sparse annotated training videos to perform temporal pose propagation and estimation. Although they bring significant improvement to recent benchmarks(e.g., MS-COCO\cite{lin2014microsoft} and MPII\cite{andriluka20142d}), it's still hard to apply in practical applications due to the high-cost annotations. In such situation, domain adaptation offers an appealing solution by adapting pose estimator from label-rich source domain to the unlabeled or few labeled target domain.

\textbf{Domain Adaptation.}
Domain adaptation utilizes a labeled source domain to learn a model that performs well on an unlabeled or sparse labeled target domain\cite{zhang2018collaborative,ganin2014unsupervised,kim2020cross-domain}.
Most methods tackle UDA by minimizing the distance\cite{long2015learning} across two distributions or aligning the output in adversarial learning\cite{tzeng2017adversarial}. For instance, \cite{hoffman2016fcns} applies the adversarial strategy to align features for semantic segmentation. \cite{wu2018dcan:} applies the generator to transfer the source data to the target style for reducing the visual differences. On one hand, these methods simply align the global coarse-level features that benefit the classification task but it's hard to align the fine-grained human pose features. We thus innovatively propose a CAFA module to adapt the consistent fine-grained features across domains. On the other hand, these methods don't need consider the topological structure of the object-of-interest, which is essential for accurate pose estimation. With this derivation, we propose IHTA mechanism to exploit the human-topological relations across domains to better bridge the inter-domain discrepancy.

A plethora of SSDA works\cite{luo2017label,saito2019semi-supervised,kim2020cross-domain} have already emerged. For example, the \cite{saito2019semi-supervised} designs a min-max entropy minimization strategy to achieve better adaptation. Some works\cite{kim2020cross-domain,luo2017label} consider the intra-adaptation bias but they either leverage the labeled data to learn discriminative features like\cite{kim2020cross-domain} or minimize the entropy similarity between intra-target samples like\cite{luo2017label}. Directly applying these methods in pose estimation cannot fully explore the keypoint semantic relationship in intra-target domain. To this end, we propose to align the one-to-one keypoint heatmap vectors in target domain for avoiding the keypoint shift chaotically.
\section{Method}
\begin{figure*}[htb]
\centering
\includegraphics[width=6.85in,height=3.2in]{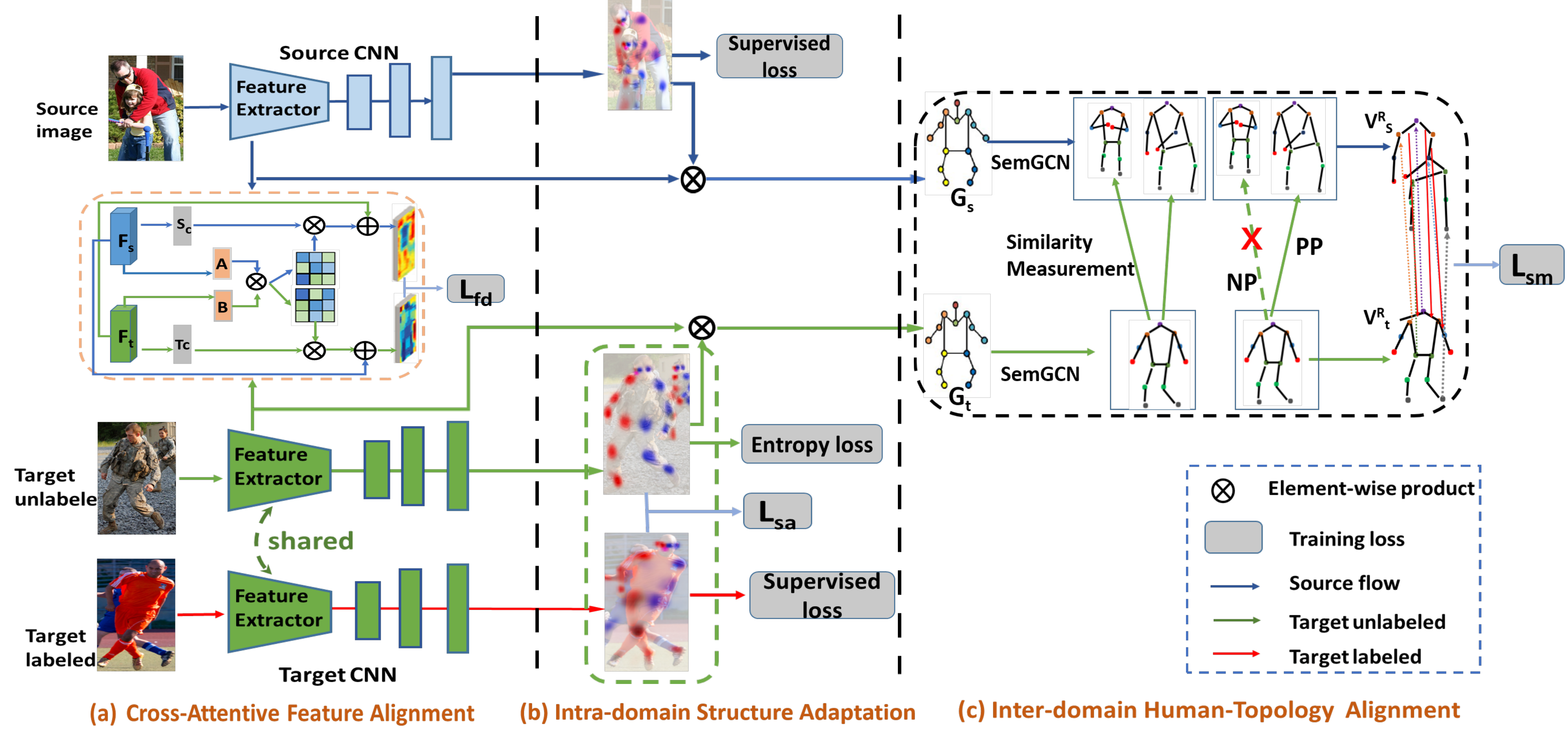}
\caption{Overview of the proposed method. The CAFA (in orange dashed line box) is placed at the encoder and learns domain-invariant fine-grained human features. The ISA (in green dashed line box) only works in SSDA setting to explore the keypoint heatmap alignment inside target domain. The IHTA (in a black dashed line box) works at the head of the network to model human pose with SemGCN\cite{zhao2019semantic} and further explore human-topology relations across domains to improve accuracy. It's noted that all the components in the dashed line box are only worked in the training phrase.}
\label{fig:2network}
\end{figure*}
\subsection{Framework Overview}
\textbf{Problem Definition.}
We denote samples in the source domain and target domain as $X_{s}$ $=$ $\{x^{i}_{s}\}^{M}_{i=1}$ and $X_{t}$ $=$ $\{x^{j}_{t}\}^{N}_{j=1}$, where $M$, $N$ are the sample numbers of each domain. Every source sample in $X_{s}$ is annotated with corresponding keypoint annotations $Y_{s}$ $=$ $\{y^{i}_{s}\}^{M}_{i=1}$. In the semi-supervised setting, we have $N_{1}$ labeled images, $N_{2}$ unlabeled images and the labeled annotations are depicted as $Y_{tl}$ $=$ $\{y^{j}_{t}\}^{N_{1}}_{j=1}$. In unsupervised setting, $N_{1} = 0$.

\textbf{Overview.} We adopt the modified SimpleBaseline\cite{xiao2018simple} as our baseline, which utilizes an encoder-decoder architecture to make pose predictions as depicted in Fig~\ref{fig:2network}. During training, given image triplets $x^{i}_{s}$, $x^{j}_{t}$ and $x^{k}_{t}$, we generate the corresponding features $f^{i}_{s}$, $f^{j}_{t}$ and $f^{k}_{t}$ with a feature extractor. Through the CAFA module, we get the adapted features $F_{s}$, $F_{t}$. Then, we feed them into the estimator and predict the respective keypoint heatmaps. In semi-supervised setting, we put the heatmap outputs into ISA module to get the aligned heatmaps. Then, we turn the heatmaps into the local keypoint features. And we formulate the human pose as graph based on them to align the topological structure via IHTA module to achieve more precise predictions.
\subsection{Cross-Attentive Feature Alignment}\label{CAFA}
The fine-grained features are effective for the accurate pose estimation. The goal of CAFA is to adapt more domain-invariant fine-grained human features across domains. Different from previous feature adaptation methods, we capture the related fine-grained feature responses across domains via our BSAM. It explores the local spatial feature dependency across domains rather than simply consider the domain features separately. The fine-grained human features can be well encoded for each domain via exploring the feature interaction in a bidirectional manner.

In specific, we design a source-to-target adaptation (STA) to enhance source human body features by adaptively aggregating the target features based on their similarity. Similarly, we also adopt target-to-source adaptation (TSA) to update target features by aggregating relevant source features. The details of CAFA is depicted in the orange dashed-box in Fig~\ref{fig:2network}.

Given the sample pairs $x_{s}$, $x_{t}$, we get the feature pairs $F_{s}$, $F_{t}$ and apply two convolution layers to generate $A$ and $B$, respectively. $F_{s}$, $F_{t}$ are also fed into another convolution layer to obtain $S_{c}$, $T_{c}$. To determine fine-grained feature dependency between each position in $F_{s}$, $F_{t}$, an correlation map $\Phi$ is formulated as $\Phi = A^{T}B$, where $\Phi^{(i,j)}$ measure the similarity between $i$-th position in $F_{s}$ and $j$-th position in $F_{t}$. To enhance $F_{s}$ with the similar response from $F_{t}$ and vice versa, the bidirectional adaptation is defined as follows.

\textbf{Source-to-Target Adaptation.}
During the STA, we define the source-to-target spatial correlated map as,
\begin{equation}
\begin{aligned}
\label{STA}
\Psi^{(i,j)}_{s \rightarrow t} = \frac{exp(\Phi^{(i,j)})}{\sum_{j=1}^{H\times W}exp(\Phi^{(i,j)})},
\end{aligned}
\end{equation}
where $\Psi^{(i,j)}_{s \rightarrow t}$ represents the impact of $i$-th position in $F_{s}$ to $j$-th position in $F_{t}$. To leverage the fine-grained features with similar spatial responses in the target domain, we update $F_{s}$ as,
\begin{equation}
\begin{aligned}
\label{feature}
F^{'}_{s} = F_{s} + \lambda_{s}T_{c}\Psi_{s \rightarrow t},
\end{aligned}
\end{equation}
where $\lambda_{s}$ leverages the importance of target-domain relevant spatial information and original source features. In this way, the target similar feature responses are well encoded in each position of $F^{'}_{s}$.

\textbf{Target-to-Source Adaptation.}
Similarly, we obtain the target-to-source attentive map $\Psi^{(i,j)}_{t \rightarrow s}$ in Eq~\ref{STA}. It indicates the impact the $j$-th position in $F_{t}$ attends to the $i$-th position in $F_{s}$. $F_{t}$ is updated by combining the similar fine-grained source-domain repsonses and original target features like in Eq~\ref{feature}. In this manner, $F^{'}_{s}$ and $F^{'}_{t}$ enable us to encode more fine-grained features for each domain.

\textbf{Loss and alignment.}
Finally, we apply the Maximum Mean Discrepancy $\textbf{(MMD)}$\cite{gretton2012a} to align $F^{'}_{s}$ and $F^{'}_{t}$ across domains in Eq~\ref{fd}. 
\begin{equation}
\begin{aligned}
\label{fd}
\mathcal{L}_{fd} =  \parallel\frac{1}{M}\sum_{i=1}^{M}\phi(F^{'}_{s,i}) - \frac{1}{N}\sum_{j=1}^{N}\phi(F^{'}_{t,j}) \parallel^{2}_{\mathcal{H}},
\end{aligned}
\end{equation}
where $\phi$ is a map operation which projects the domain into a reproducing kernel Hilbert space $\mathcal{H}$ \cite{gretton2006a}. The arbitrary distribution of features can be represented by the kernel embedding technique. It allows us to learn domain-invariant and fine-grained human representations across domains by minimizing $\mathcal{L}_{fd}$.
\subsection{Intra-domain Structure Adaptation}
\label{ISA}
Under the SSDA setting, the $\textbf{T}$arget $\textbf{L}$abeled \textbf{(TL)} and $\textbf{T}$arget $\textbf{U}$nlabe
led \textbf{(TU)} data have a quite potential relationship actually. On one hand, the scales, postures, or appearances of people are varied between them. On the other hand, they subject to a homogeneous distribution and possess similar specific keypoint information. Obviously, the TL is much amenable to acquire more accurate predictions than TU because it contains the detailed annotations. We hypothesize that our model can discover the underlying one-to-one keypoint semantic correspondence across them which benefits recognizing the vague keypoint locations of TU. The TL can provide more explicit guidance to facilitate TU to better rectify the inaccurate localizations (e.g., the confused keypoint locations of baseline with ISA is more explicit in Fig~\ref{fig:isa}). Concretely, we devise ISA module to encourage TL to augment the corresponding keypoint representations of TU via calculating the cosine similarity of their heatmap vectors $y'^{j}_{t}$ and $y'^{k}_{t}$ as follows,
\begin{equation}
\begin{aligned}
\label{ISA_loss}
\mathcal{L}_{sa} =\sum_{j=1}^{N_{1}}\sum_{k=1}^{N_{2}} \sum_{h=1}^{H}\frac{y'^{j}_{t}(h) \cdot y'^{k}_{t}(h)}{\parallel y'^{j}_{t}(h)\parallel\cdot\parallel y'^{k}_{t}(h) \parallel},
\end{aligned}
\end{equation}
where $H$ is the keypoint number and the $\mathcal{L}_{sa}$ measures the keypoint heatmap similarity of the same category between TL and TU. We align them to force TL to guide TU with the exclusive semantic prior via minimizing the $\mathcal{L}_{sa}$.
\subsection{Inter-domain Human-Topology Alignment}
\label{IHTA}
Although we align the one-order keypoint features of intra-domain, it cannot better conquer large pose discrepancy with severe geometric deformation, especially in heavily occluded cases across domains. However, the domain-invariant human-topology knowledge can provide a reliable guidance for mitigating this issue. We thus propose an IHTA mechanism to preserve this information. Since the human body structure provides the essential constraint information between joints, our IHTA is designed by GCN, which offers an explicit way of modeling the high-order human skeleton structure that is advantageous for capturing the spatial topological information of joints. It makes the cross-domain human-topology adaptation effective and conceivable. The details are as below.   

\textbf{Local Keypoint Feature Extraction.}
Specifically, we can get a group of semantic local features of keypoints $V^{kp}$ according to the above feature maps $F$ and keypoint heatmaps $y^{kp}_{i}$ for both domains via an outer product($\otimes$) and a global average pooling in Eq~\ref{heatmapp}.
\begin{equation}
\begin{aligned}
\label{heatmapp}
V^{kp} = \{v_{i}\}_{i=1}^{H} = Global(F \otimes y^{kp}_{i}).
\end{aligned}
\end{equation}

\textbf{Graph Formulation.} Here, we construct an intuitive graph $G = (V,E)$  based on the keypoint local features in Eq~\ref{heatmapp} for each human pose. $V$ is the node set in $G$ which can be denoted as $V = \{{v_{i}, i = 1, 2, ..., H}\}$. $E =$ $\{$${v_{i}v_{j}}$ $|$ if $i$ and $j$ are connected in the human body$\}$ is the edge set which refers to limbs of the human body. The adjacent matrix of $G$ refers to matrix $A = {a_{ij}}$, with $a_{ij} = 1$ when $v_{i}$ and $v_{j}$ are neighbors in $G$ or $i = j$, otherwise $a_{ij} = 0$. 

\textbf{Graph Convolution Network.} Our key insight is that human body structure is a natural graph and there exists potential spatial constraint among joints. Hence, we model the human-topology representation via the recent SemGCN\cite{zhao2019semantic}. For a graph convolution, propagating features through neighbor joints helps to learn robust local structure and relation information between joints. Meanwhile, the non-local layer\cite{wang2018non-local} helps capture the local and global long-range dependency among nodes to learn more human context information. It enables us to harvest robust human-topology informations, which are essential for learning structure-invariant information across domains. 

A graph based convolutional propagation applies to node $i$ in two steps. Firstly, node representations are transformed by a learnable parameter matrix $W \in R^{D_{l+1}\times D_{l}}$. Second, the transformed node representations are gathered to node $i$ from its neighboring node $j$, followed by a RELU function. The node features are collected into a matrix ${v}^{(l)} \in R^{D_{l}\times H}$. Following the SemGCN\cite{zhao2019semantic}, a different weighting matrix is applied to each channel $d$ of node features:
\begin{equation}
\begin{aligned}
\label{graph}
v^{(l+1)} = \Arrowvert^{D_{l+1}}_{d=1}\sigma(\vec{w}_{d}v^{(l)}\varphi_{i}(M_{d}\odot A)),
\end{aligned}
\end{equation}
where $v^{(l)} \in R^{D_{l}}$ and $v^{(l+1)}\in R^{D_{l+1}}$ are the node representations before and after $l$-th convolution respectively, $M_{d}$ is a set of $M \in R^{H \times H}$, which is a learnable weighting matrix compared with vanilla graph convolution. The weight vectors show the local semantic knowledge of neighboring joints implied in the graph. The $\Arrowvert$ depicts channel-wise concatenation, and $\vec{w}_{d}$ is the $d$-th row of the transformation matrix $W$.
It learns channel-wise weights for edges as priors in the graph (e.g., how one joint influences other body parts in pose estimation) to enhance the graph representations. And $\varphi_{i}$ is Softmax nonlinearity which normalizes the input matrix across all choices of node $i$, $\odot$ is an element-wise operation which returns $m_{ij}$ if $a_{ij} = 1$. $A$ forces that for node $i$, we only compute the weights of its neighboring nodes $j$. Hence, the relationship of neighboring nodes are well considered.

Then, aside by the non-local concept\cite{wang2018non-local} and we define the feature updating operation as:
\begin{equation}
\begin{aligned}
\label{graph1}
v^{(l+1)}_{i} = v^{(l)}_{i}+\frac{W_{v}}{H}\sum_{j=1}^{H}f(v^{(l)}_{i},v^{(l)}_{j})\cdot g(v^{(l)}_{j}),
\end{aligned}
\end{equation}
where $W_{v}$ is initialized as zero; $f$ is to compute the affinity between node $i$ and all other $j$; $g$ computes the node $j$ representation. It computes responses between joints with their features to capture local and global long-range relationships among nodes. 

\textbf{Sample Selection Mechanism.} It's unnecessary to conduct a hard alignment for the arbitrary poses in two domains since any two poses have the different shapes and geometric representations originally. Thus, we should predefine the 'similarity standard' for choosing the pose pairs should be aligned across domains to avoid nonsense alignment. Firstly, we calculate the averaged keypoint features similarity between two poses across domains $\Gamma_{sim}$ in Eq~\ref{pp},
\begin{equation}
\begin{aligned}
\label{pp}
\Gamma_{sim} = \frac{1}{H}\sum_{i=1}^{H}\sqrt{c^{s}_{i}c^{t}_{i'}}|v^{s}_{i}-v^{t}_{i'}|,
\end{aligned}
\end{equation}
where the $c^{s}_{i}$ and $v^{s}_{i}$ indicate the confidence value and keypoint features of $i$-th keypoint of a source sample and the $i'$ is the corresponding one in target domain. Then, we define a threshold value $\tau$ to judge whether the human pairs $(m,n)$ are similar. As shown in Eq~\ref{pair}, if the similarity is above or equals to $\tau$, we view them as positive pairs (PP) otherwise are negative pairs (NP). The value of $\tau$ is $0.7$ here.
\begin{equation}
    Pair_{(m,n)}\in
   \begin{cases}
   \label{pair}
   PP&\mbox{if $\Gamma_{sim}$ $>=$ $\tau$ }\\
   NP&\mbox{else}.
   \end{cases}
\end{equation}

\textbf{Cross-Graph Topology-Alignment.} Based on the above steps, we conduct a cross-graph alignment to align the joint relation information learned by SemGCN\cite{zhao2019semantic} across two humans. Rather, given two samples $x_{s}$ and $x_{t}$ from both domains respectively, we firstly obtain the updated joint representations via Eq~\ref{graph},~\ref{graph1}. And followed by a $1 \times 1$ convolution, we get the final human-topology representations $V^{R}_{s}$, $V^{R}_{t}$. Finally, we choose the positive pairs via Eq~\ref{pair} and align their $V^{R}_{s}$, $V^{R}_{t}$ via Eq~\ref{Tdistance}.
\begin{equation}
\begin{aligned}
\label{Tdistance}
\mathcal{L}_{sm} = \sum_{i\in M}\sum_{k \in N_{2}}cosine(V^{R}_{s_{i}},V^{R}_{t_{k}}),
\end{aligned}
\end{equation}
where $V^{R}_{s_{i}}$ and $V^{R}_{t_{k}}$ indicate the topological representation of the $i$-th sample for each domain. We can learn the generalized high-order structure-invariant representations on both domains by minimizing $\mathcal{L}_{sm}$ to help reason the keypoint locations.
\subsection{Optimization}
\label{optimization}
The training of our network is to minimize a weighted combination of the aforementioned loss with respect to their parameter:
\begin{equation}
\begin{aligned}
\mathcal{L}_{pose} = \beta_{sup} \mathcal{L}^{sup}_{pose} + \beta_{da} \mathcal{L}^{da}_{pose} ,
\end{aligned}
\end{equation}
where the weights $\beta_{sup}$ and $\beta_{da}$ are chosen empirically to strike a balance among the model capacity, and prediction accuracy. 

\textbf{Supervised Pose Loss.}
The supervised loss consists of the $L^{s}_{pose}$ for labeled source data and the $L^{tl}_{pose}$ for the TL prediction. The mean square error (MSE) is adopted as our regression loss:
\begin{equation}
\begin{aligned}
\mathcal{L}^{s}_{pose} =  \sum_{i=1}^{M}\sum_{h=1}^{H}\parallel y^{h}_{s_{i}} - y'^{h}_{s_{i}} \parallel^{2}_{2},
\end{aligned}
\end{equation}
\begin{equation}
\begin{aligned}
\label{TL_loss}
\mathcal{L}^{tl}_{pose} = \sum_{j=1}^{N_{1}} \sum_{h=1}^{H}\parallel y^{h}_{t_{j}} - y'^{h}_{t_{j}} \parallel^{2}_{2},
\end{aligned}
\end{equation}
\begin{equation}
\begin{aligned}
\mathcal{L}^{sup}_{pose} =  \alpha_{tl} \mathcal{L}^{tl}_{pose} + \mathcal{L}^{s}_{pose},
\end{aligned}
\end{equation}
where $\alpha_{tl}$ is the trade-off hyperparameter and its value is $0.5$, the Eq~\ref{TL_loss} is only used in semi-supervised setting.

\textbf{Entropy Loss.}
Entropy minimization (ENT)\cite{grandvalet2004semi-supervised} is a semi-superv
ised method assuming that the model is confident about its prediction for the unlabeled data. We adopt it as a regularizer and ensure that it maximally helps the TU achieve better performance in target domain. We add this term to the optimization in Eq~\ref{ISA_loss}:
\begin{equation}
\begin{aligned}
\mathcal{L}^{tu}_{ent} =  \sum_{k=1}^{N_{2}} ent(y'^{k}_{t}),
\end{aligned}
\end{equation}
where $ent (p)$ calculates the entropy of distribution $p$.

\textbf{Domain Adaptation Loss.} In this case, the domain adaptation loss consists of the above three adaptation losses as follows:
\begin{equation}
\begin{aligned}
\mathcal{L}^{da}_{pose} = \alpha_{sa} \mathcal{L}_{sa} + \alpha_{sm}\mathcal{L}_{sm} + \alpha_{fd} \mathcal{L}_{fd},
\end{aligned}
\end{equation}
where the $\alpha_{sa}$, $\alpha_{sm}$, $\alpha_{fd}$ are the weighted balance factors to keep the model effective and their values will be discussed in section~\ref{loss_weight}. 
\section{Experiments}
\subsection{Datasets and Evaluation metric}
The proposed method is evaluated on two recent multi-person datasets: MPII Human Pose benchmark\cite{andriluka20142d}, COCO 2017 Keypoints Detection dataset\cite{lin2014microsoft}. 

\textbf{COCO Keypoint Detection} consists of the training set (includes $57$$K$ images),
the test-dev set (includes around $20$$K$ images) and the validation set (includes $5$$K$ images). The MS-COCO evaluation metrics, OKS-based average precision (AP) and average
recall (AR), are used to evaluate the performance. The OKS (object keypoints similarity) defines the similarity between the predicted heatmap and the groundtruth.

\textbf{MPII Human Pose Dataset} consists of images taken from real-world activities with full-body pose annotations. There are about $25$$K$ images with $40$$K$ objects, where there are $12$$K$ objects for testing and the remaining for the training set. We use the standard metric PCKh\cite{andriluka20142d} (head-normalized probability of correct keypoint) score as evaluation. The PCKh@$0.5$ score is reported in our results, $50$\% of the head size for normalization.

In our experiment, there are two source-target settings:

1. Source: MS-COCO/ Target: MPII.

2. Source: MPII/ Target: MS-COCO.
\subsection{Implementation Details}
\textbf {Network architectures.} We adopt SimpleBaseline\cite{xiao2018simple} and HRNet\cite{sun2019deep} as the pose estimation baseline respectively and the backbone uses the ResNet-$50$ in default. As for the SemGCN\cite{zhao2019semantic}, the building block is one residual block\cite{he2016deep} built by two SemGConv layers with $128$ channels, followed by one non-local layer\cite{wang2018non-local}. This block is repeated four times. All SemGConv layers are followed by Batch Normalization and a RELU activation except the last one.

\textbf {Data Augmentation.} We apply random flip, rotation, and scale in our training stage. The flip value is $0.5$. The scale range is ([$0.7$ $\sim$ $1.3$]), and the random rotation range is ([$-40$$^{\circ}$C $\sim$ $+40$$^{\circ}$C]).

\textbf {Training.} We implement all the experiment in PyTorch\cite{Paszke2017AutomaticDI} on a single NVIDIA TITAN XP GPU with $12$ GB memory. We select the ResNet-$50$, $101$, $152$ as the backbones which are all initialized with the weights of the ImageNet\cite{russakovsky2015imagenet} pretrained model. We use Adam optimizer\cite{kingma2015adam:} with learning rate $10^{-4}$, momentum $0.9$ and weight decay $10^{-4}$ to train our model. We train the model for $150$ epochs taking $96$ hours for MS-COCO and we resize the images to $256$$\times$$256$ and trained for $135$ epochs taking $10$ hours for MPII.

It's noted that we don't evaluate the results on MS-COCO test-dev dataset due to it need to test online and unequal keypoint numbers across dataset makes it harder. We unify the MS-COCO keypoint numbers with the MPII finally.
\subsection{Domain Adaptation Performance}
We compare the performance of our method with the supervised approaches on both domains. For comparison, we take the Direct Transfer (DT) that trained with the source only as the baseline. To further shed light on the effectiveness of our method, we list the results of the model trained without the annotations absolutely (UDA model) and with the few labeled data in target domain (Adaptation model) respectively. Our method brings competitive performance with either data setting.

\textbf{MPII to MS-COCO.} 
The table~\ref{tab:1 COCO minival} reports the results on MPII to MS-COCO. Although the scale of the source domain is much less than the target and the complexity and difficulty are also inferior to it, the UDA model can still achieve $64.4\%$ AP, which is higher $7.7\%$ than baseline although the target domain without any supervision. It powerfully proves that our method is conductive to weaken the influence from the irrelevant source information thereby encouraging relevant knowledge transfer among shared representations even without label. Although we adopt only $40\%$ target labeled data, it works effectively outperforming UDA by $2.1\%$, indicating that ISA further alleviates the intra-domain bias. Moreover, the accuracy also improves despite we adopt HRNet\cite{sun2019deep} as the baseline model, which further illustrates our method with pretty generality.
\begin{table}[htp]
\centering
\setlength{\tabcolsep}{0.1mm}
\caption{The comparison results on MPII to MS-COCO}
\label{tab:1 COCO minival}
\begin{tabular}{c|c|ccccccc}
    \toprule
    Method & Backbone &  AP & AP.5 & AP.75 & AP(M) & AP(L) & AR  \\
     \midrule
     \multicolumn{7}{c}{\textbf{Supervised methods (Target only)}} \\
     \hline
  G-RMI\cite{papandreou2017towards}& - & 65.7& 83.1 & 72.1& 61.7 & 72.5&69.9\\
  MultiPoseNet\cite{kocabas2018multiposenet:} &  -   &69.6& 86.3 & 76.6  & 65.0 & 76.3 &  73.5\\
  AE\cite{newell2017associative}&Hourglass&66.3&86.5&72.7&61.3& 73.2&71.5\\
  CSANet\cite{2019Pose}&ResNet-50 & 72.1&-& - &- &- & - \\
  SBN\cite{xiao2018simple} &ResNet-50 & 70.4&88.6& 78.3& 67.1& 77.2&  76.3 \\
  HRNet\cite{sun2019deep} & HRNet-W32 & 74.4&90.5& 81.9& 70.8& 81.0&  79.8 \\
  HRNet\cite{sun2019deep} & HRNet-w48   & 75.1&90.6& 82.2& 72.5& 81.8&  80.4 \\
     \hline
      \multicolumn{7}{c}{\textbf{Direct Transfer (Source only)}} \\
      \hline
    SBN & ResNet-50 & 56.7 & 72.3 & 64.4 & 54.3 & 53.4&  62.2  \\
    HRNet & HRNet-w32 &  59.1 & 75.3  & 66.7 & 55.4 & 66.3&  64.3 \\
    HRNet & HRNet-w48 &  59.5 &75.2 & 66.8 & 55.6& 66.8 &64.9  \\
    \hline
    \multicolumn{7}{c}{\textbf{UDA model (Without ISA module)}} \\
      \hline
    SBN & ResNet-50  & 64.4 & 80.4 &71.2  & 62.3 & 71.4&  70.4  \\
    HRNet & HRNet-w32  & 67.2  & 83.3  & 74.6 & 63.4 & 74.3& 72.7 \\
    HRNet & HRNet-w48  & 67.8  & 83.2   & 74.3 & 63.6& 74.6 &  73.4 \\
    \hline
     \multicolumn{7}{c}{\textbf{Adaptation model (Ours)}} \\
     \hline
    Ours(SBN) & ResNet-50  & \textcolor[rgb]{1.00,0.00,0.00}{66.5} & 82.3 & 73.2 & 64.1 & 73.5&  72.3  \\
    Ours(HRNet) & HRNet-w32 &\textcolor[rgb]{1.00,0.00,0.00}{69.2}  & 85.2  & 76.5 &65.6 & 76.1 &74.6  \\
    Ours(HRNet) & HRNet-w48&\textcolor[rgb]{1.00,0.00,0.00}{69.9}  &85.3 &76.4  &65.7 & 76.5 &75.1  \\
     \bottomrule
\end{tabular}
\end{table}

\begin{figure}[htp]
\centering
\includegraphics[width=3.2in,height = 2.0in]{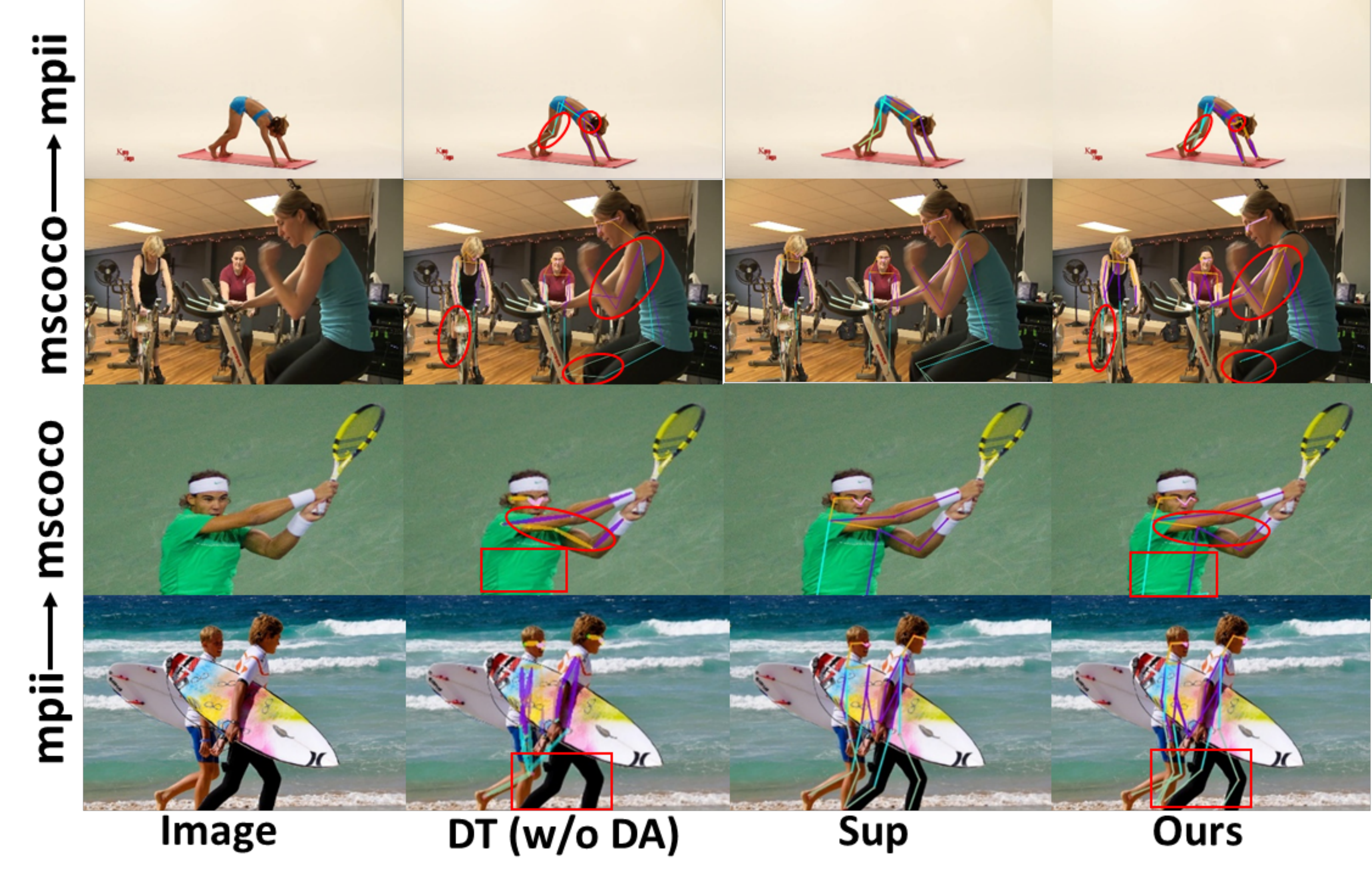}
\caption{Results comparison of the baseline (DT (w/o DA)) versus the supervised model (Sup) and Ours on both cases.}
\label{fig:3mpii2coco}
\end{figure}

\begin{figure}[htp]
\centering
\includegraphics[width=3.0in,height = 1.8in]{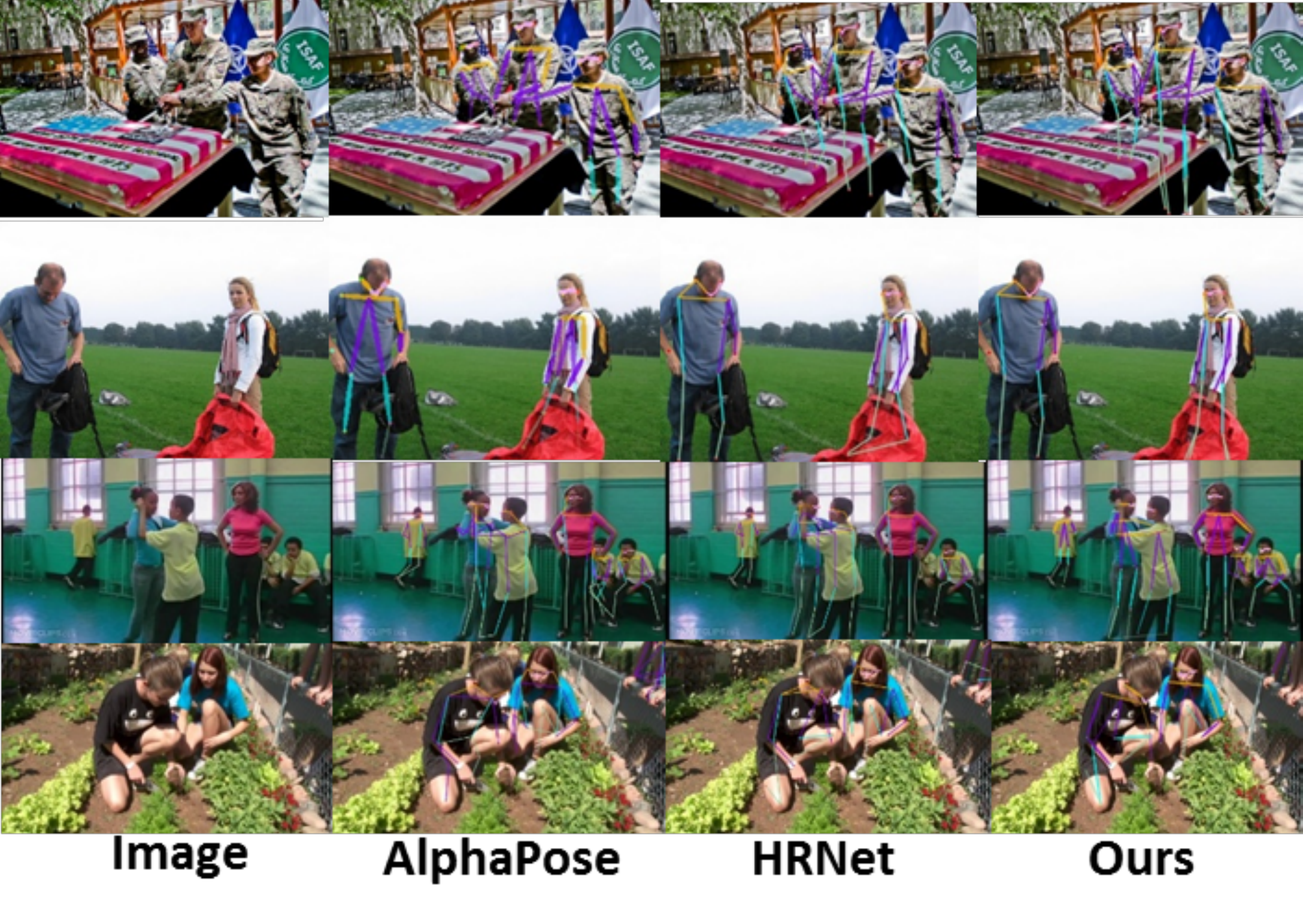}
\caption{The visual comparisons between existing methods (they are under fully-supervised setting and without adaptation) and Ours, the samples are randomly selected from the MPII/MS-COCO.}
\label{fig:comparison}
\end{figure}

The qualitative visualization is shown in Fig~\ref{fig:3mpii2coco}. It shows that our method achieves impressive results compared with the baseline, and achieves a comparable performance with the supervised model (e.g., the explicit predictions highlighted with the red circle/box). We also did visual comparisons with other recent pose estimation methods as shown in Fig~\ref{fig:comparison}, it can further manifest the robustness of our method.

\textbf{MS-COCO to MPII.}
We evaluate the PCKh@$0.5$ score on MSCOC
O to MPII in Tab~\ref{tab:2 MPII} for comparison. The UDA model achieves $85.2\%$ PCKh@$0.5$ and improves $7.9\%$ points without annotations than the baseline. It indicates that minimizing the domain distribution difference is essential for tackling pose domain shift. Additionally, $40\%$ labeled data are adopted for target domain (Ours) can further improves PKCh@$0.5$ to $87.8$, which outperforms the baseline $10.5\%$. It proves that our method ensures reliable knowledge adaptation in different domains and receives the decent localized results. We also test on the HRNet-w$32$ and obtain the best result $89.4$.
\begin{table}[ht]
\caption{The comparison results on MS-COCO to MPII}
\label{tab:2 MPII}
\centering
\setlength{\tabcolsep}{0.10mm}
\begin{tabular}{c|ccccccc|c}
\toprule
    Method & Head & Shoulder & Elbow & Wrist & Hip & Knee & Ankle & Total \\
    \midrule
     \multicolumn{9}{c}{\textbf{Supervised methods (Target only)}} \\
   \hline
    Wei $\emph{et al.}$\cite{wei2016convolutional} & 97.8 & 95.0 & 88.7 & 84.0 & 88.4 & 82.8 & 79.4 & 88.5 \\
    Newell $\emph{et al.}$\cite{newell2016stacked} & 98.2 & 96.3& 91.2 & 87.2 & 89.8 & 87.4 & 83.6 & 90.9 \\
    Sun $\emph{et al.}$\cite{sun2017human} & 98.1 & 96.2 & 91.2& 87.2 & 89.8 & 87.4 & 84.1 & 91.0 \\
    Tang $\emph{et al.}$\cite{tang2018quantized} & 97.4 & 96.4 & 92.1 & 87.7 & 90.2 & 87.7 & 84.3 & 91.2 \\
    Ning $\emph{et al.}$\cite{ning2018knowledge-guided} & 98.1 & 96.3 & 92.2 & 87.8 & 90.6 & 87.6 & 82.7 & 91.2 \\
    Luvizon $\emph{et al.}$\cite{luvizon2017human} & 98.1 & 96.6 & 92.0 & 87.5 & 90.6 & 88.0 & 82.7 & 91.2 \\
    Chu   $\emph{et al.}$\cite{chu2017multi-context} & 98.5& 96.3 & 91.9 & 88.1 & 90.6 & 88.0 & 85.0 & 91.5 \\
    Chou $\emph{et al.}$\cite{chou2018self} & 98.2& 96.8 & 92.2 & 88.0 & 91.3 & 89.1 & 84.9 & 91.8 \\
    Chen $\emph{et al.}$\cite{chen2017adversarial} & 98.1 & 96.5 & 92.5 & 88.5 & 90.2 & 89.6 & 86.0 & 91.9 \\
    Yang $\emph{et al.}$\cite{yang2017learning} & 98.5 & 96.7 & 92.5 & 88.7 & 91.1 & 88.6 & 86.0 & 92.0 \\
    Ke $\emph{et al.}$\cite{ke2018multi-scale} & 98.5 & 96.8 & 92.7 & 88.4 & 90.6 & 89.3 & 86.3 & 92.1 \\
    Tang $\emph{et al.}$\cite{tang2018deeply} & 98.4 & 96.9 & 92.6 & 88.7 & 91.8 & 89.4 & 86.2 & 92.3 \\
   \hline
    SBN $\emph{et al.}$\cite{xiao2018simple} & 98.5 & 96.6 & 91.9 & 87.6 & 91.1 & 88.1 & 84.1 & 91.5 \\
    HRNet $\emph{et al.}$\cite{sun2019deep} & 98.6 & 96.9 & 92.8 & 89.0 & 91.5 &89.0 & 85.7 & 92.3 \\
     \hline
      \multicolumn{9}{c}{\textbf{Direct Transfer (Source only)}} \\
     \hline
    SBN-ResNet50& 95.2  & 89.8 &79.7  &72.5  &75.8  & 67.8 &60.5  & 77.3 \\
    HRNet-w32& 95.8 & 90.3 & 80.5 & 74.3 & 77.6 & 69.7 & 62.8 & 79.6 \\
    \hline
     \multicolumn{9}{c}{\textbf{UDA model (Without ISA module)}} \\
     \hline
    SBN-ResNet50& 96.8  &96.5  & 86.1 &81.0  &86.4 &80.2  & 73.1 & 85.2 \\
    HRNet-w32&95.8  & 94.5 & 87.8 & 83.5 & 87.5 & 82.5 & 79.1 & 86.4 \\
    \hline
    \multicolumn{9}{c}{\textbf{Adaptation model (Ours)}} \\
    \hline
     Ours(SBN)& 97.5  & 94.4 &86.8  &82.2  &87.2  & 81.2 &73.9  & \textcolor[rgb]{1.00,0.00,0.00}{87.8} \\
    Ours(HRNet-w32)& 97.2 & 95.6 & 89.8 & 84.6 & 88.7 & 83.8 & 80.2 & \textcolor[rgb]{1.00,0.00,0.00}{89.4} \\
    \bottomrule
  \end{tabular}
   \end{table}
   
As shown in Fig~\ref{fig:3mpii2coco}, our approach makes more accurate predictions than DT and its supervised model ($\emph{e.g.}$, the localized joints highlighted with red box) and also achieves a competitive result with the other popular estimation methods like in Fig~\ref{fig:comparison}. 
\subsection{Ablation Study for Design Modules}
To analyse the effectiveness of each component, we conduct experiments to evaluate their contributions. The SimpleBaseline\cite{xiao2018simple} (ResNet-$50$) is adopted for our default baseline. The ablated results are illustrated in Table~\ref{tab:3 DT}, the SL depicts model that both trained and tested on the target domain. DT is stated as above. Ours contains all components.

\textbf{Does our method really learn fine-grained and domain-invariant human features?}
Table~\ref{tab:3 DT} shows that CAFA delivers a large performance gain than DT in all cases. The result on MPII to MS-COCO achieves $63.4\%$ AP outperforming DT by $6.7\%$ solely with CAFA. The similar improvements can also be observed on MS-COCO to MPII. The PKCh@$0.5$ score grows to $84.5$, which achieves $7.2\%$ higher than DT. This indicates mitigating the feature discrepancy across domains is crucial for addressing domain shift. We also compare CAFA with the other feature adaptation stretagies. The result demonstrates our CAFA achieves outstanding performance even compared with the recent strong alignment strategy\cite{saito2019strong-weak}.
\begin{table}[h]
\centering
\setlength{\tabcolsep}{0.1mm}
\caption{Component comparisons using MPII / MS-COCO as the source dataset and MS-COCO / MPII as the target dataset.}
\label{tab:3 DT}
\begin{tabular}{ccc}
      \toprule
    Methods & MS-COCO$\rightarrow$ MPII & MPII$\rightarrow$ MS-COCO  \\
        \midrule
     \specialrule{0.05em}{-0.5pt}{0.5pt}
    SL(Trained on Target) &  $\textbf{91.5}$ & $\textbf{70.4}$ \\
      \hline
   DT(Trained on Source) & 77.3 & 56.7 \\
    \hline
    DT+DAN\cite{long2015learning}& 79.1  & 57.2 \\
    DT+Adversial DA\cite{ganin2016domain-adversarial} & 79.5  & 58.1 \\
    DT+SW Detection\cite{saito2019strong-weak} &81.6  & 60.1 \\
      DT+CAFA& 84.5 & 63.4 \\
     \hline
    DT+ISA& 83.6 &62.5 \\
  DT+IHTA& 85.6 & 64.8 \\
     \hline
    $\textbf{Ours}$  & \textcolor[rgb]{1.00,0.00,0.00}{87.8} & \textcolor[rgb]{1.00,0.00,0.00}{66.5}  \\
    \bottomrule
  \end{tabular}
\end{table}

\begin{figure}[htp]
\centering
\includegraphics[width=2.9in,height = 1.6in]{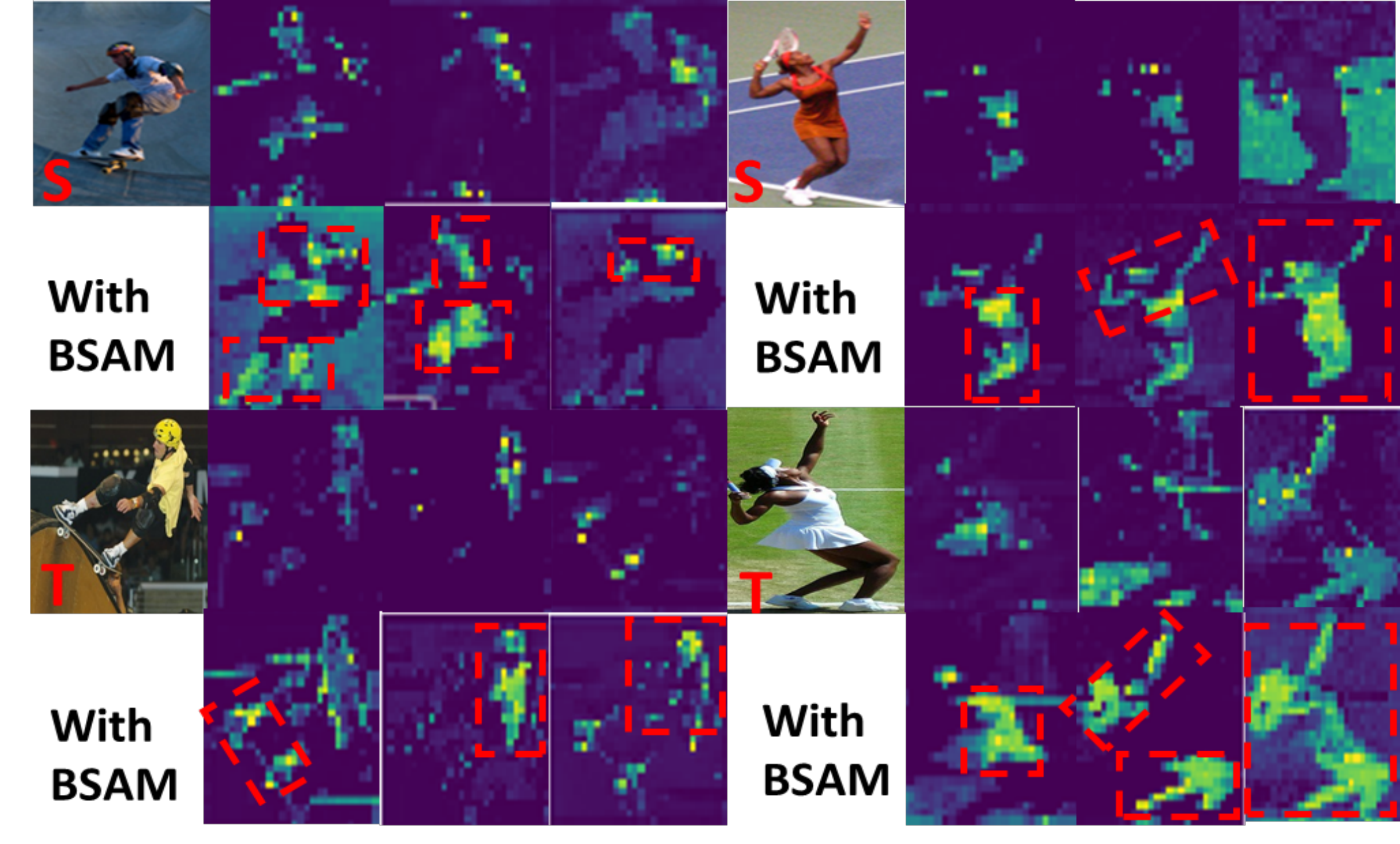}
\caption{Visualization of features with or without (baseline) our BSAM, best viewed in color.}
\label{fig:BSAM}
\end{figure}

To illustrate CAFA can adapt the fine-grained human features via BSAM mechanism, we visualize the feature maps on both domains in Fig~\ref{fig:BSAM}. As shown in red-dashed boxes, many feature maps of baseline are confused or noisy, e.g., the human part regions are ambiguous, or irrelevant regions are activated. On the contrary, our BSAM can effectively discover the similar fine-grained body parts features for each domain and enhance the respective features.

\begin{figure}[htp]
\centering
\includegraphics[width=3.4in,height = 0.86in]{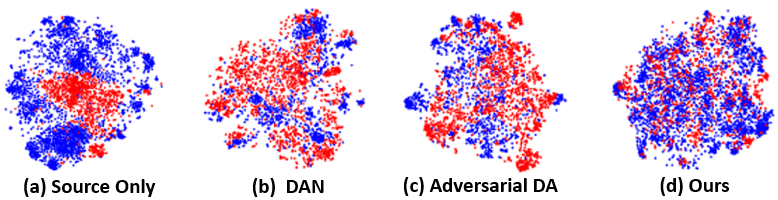}
\caption{Visualization of features using t-SNE: (a) the baseline (b) DAN (c) adversarial DA (d) Ours. Note that the blue and red points are samples from the source and target domain respectively, best viewed in color.}
\label{fig:feature}
\end{figure}

To further manifest that CAFA can produce domain-invariant features. We visualize the features of MS-COCO to MPII learned from ResNet, DAN, Adversarial DA and Ours respectively with t-SNE\cite{dermaaten2008visualizing} in Figure~\ref{fig:feature} (a)-(d). From left (ResNet) to right (Ours), the source and target domains become more and more indistinguishable. Firstly, the features of "ResNet-$50$" are not well aligned in both domains. For "DAN"\cite{long2015learning}, two domains are aligned somewhat, however, the structure of target features is scattered and the shared features are not well aligned. As for "adversarial DA"\cite{ganin2016domain-adversarial}, the target features are well preserved, but the shared features are not well aligned. For ours, the shared feature structure are compact and better aligned while the instance features of each domain are indistinguishable, which clearly evidences CAFA well captures domain-invariant features.

\textbf{The effectiveness of the ISA.}
To better elaborate that the similar pose representations from the labeled data can provide a reliable semantic guidance for the unlabeled, we conduct experiment reporting the result in Tab~\ref{tab:3 DT}. It verifies our model with ISA achieves substantial improvements over the baseline in two adaptation cases. Concretely, the accuracy increases from $77.3\%$ to $83.6\%$ directly on MS-COCO to MPII and $56.7\%$ to $62.5\%$ on MPII to MS-COCO.

\begin{figure}[htp]
\centering
\includegraphics[width=3.0in,height = 0.95in]{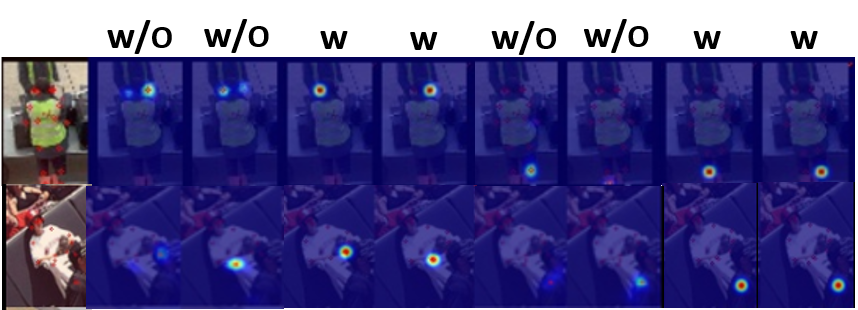}
\caption{The keypoint heatmap predictions visualization. The 'w/o' depicts the results of DT, 'w' depicts with ISA.}
\label{fig:isa}
\end{figure}

As shown in Fig~\ref{fig:isa}, we notice that the keypoint locations produced by DT are ambiguous and inaccurate. By contrast, the refined heatmaps with the ISA are more accurate and explicit. The ISA can correct subtle localized confusions of unlabeled data aided by the keypoint structural information learned from the labeled data. It exactly demonstrates our ISA well diminishes the intra-domain aliasing.

\textbf{Analysis of proposed IHTA.}
Table~\ref{tab:3 DT} shows that DT model behaves unsatisfying and only achieves poor result. The result with IHTA is improved by $8.3\%$ and $8.1\%$ on the MS-COCO to MPII and MPII to MS-COCO respectively. This obviously proves the effectiveness of the IHTA. Another innovation of IHTA is to formulate the graph model capturing the spatial relation information between joints. The result indirectly proves that adopting the graph model is robust and effective. Compared with the other components, IHTA makes the greatest contribution. It demonstrates the significance of exploring the high-order topological relation across domains.

\begin{figure}[htp]
\centering
\includegraphics[width=3.1in,height = 0.8in]{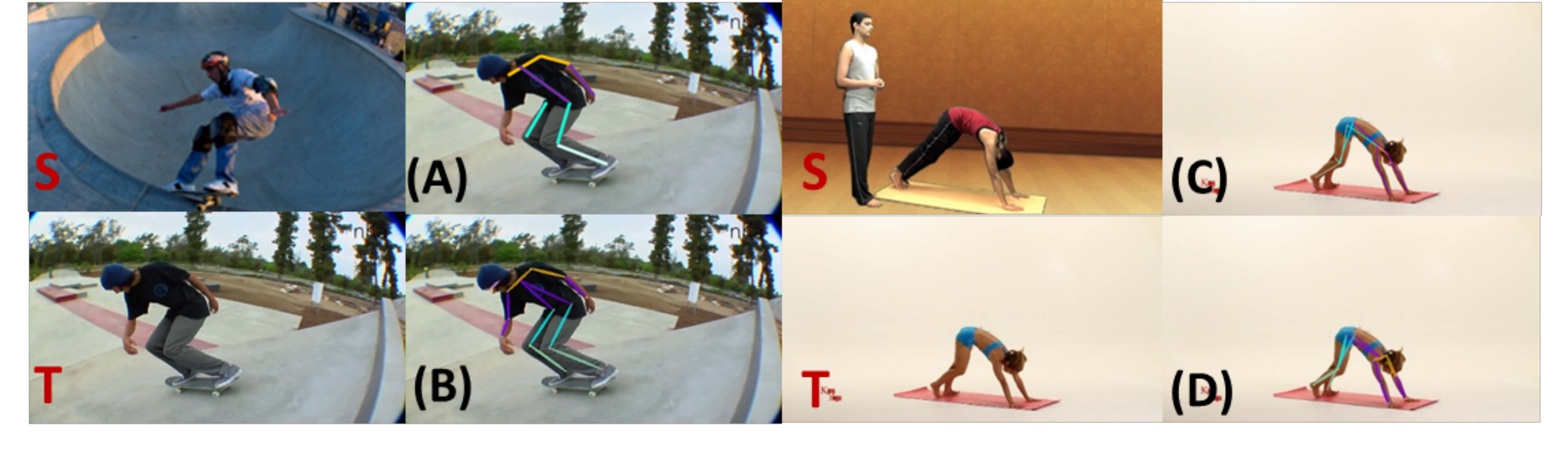}
\caption{The qualitative comparisons of IHTA. "S" and "T" refer to the source and target image. (A)/(C) represent results generated by DT model. (B)/(D) depict results with IHTA.}
\label{fig:ihtm}
\end{figure}

We also visualize the comparison results in Fig~\ref{fig:ihtm}, we observe that IHTA can fix some challenging structural errors (e.g., the occluded right ankle in Fig~\ref{fig:ihtm}(A) and the left hip/shoulder in Fig~\ref{fig:ihtm}(C)) via the human topology knowledge learned from the source domain.

\textbf{Exploring the proper value of the objective weights.}
\label{loss_weight}
Firstly, we analyze the proper value of the objective weights $\alpha$, $\beta$ in table~\ref{tab:param}. We investigate their value by varying in ${\{0.3 \sim 0.9\}}$. Besides, we testify the result on MPII to MS-COCO and similar robust behavior can also be verified on MS-COCO to MPII. For $\alpha$, as it varies from $0$ to $0.5$, the prediction accuracy on MS-COCO increases. It is desirable that when MS-COCO dataset is well-aligned with the MPII, while increasing $\alpha$ much, it preserves more general knowledge leading to negative transfer.
\begin{table}[hpt]
\centering
\caption{Sensitivity of mAP to the hyper-parameter $\alpha$, $\beta$ on MPII to MS-COCO.}
\label{tab:param}
\begin{tabular}{c|ccccccc}
  \toprule
    $\alpha$|$\beta$&0.3&0.4&0.5&0.6&0.7&0.8&0.9 \\
     \midrule
     mAP( $\alpha$) & 60.0&61.2& \textbf{64.8}&64.1& 63.5&62.2& 61.5 \\
      mAP( $\beta$)& 58.4&62.3& \textbf{64.8}&63.8& 63.2&62.7& 62.0\\
     \bottomrule
  \end{tabular}
   \end{table}
   
As for $\beta$, the accuracy also shows a growing trend as the value changes from $0.3$ to $0.5$. It illustrates that our adaptation method benefits adaptation performance improvement. When it's beyond $0.5$, it means overemphasizing the contribution of the supervised optimization leads to insufficient adaptation, which results in performance drop. This shows that proper trade-off will enhance effective knowledge transfer across domains.
\section{Conclusion}
In this paper, we propose a novel domain adaptation framework to address the multi-person pose estimation problem. In the feature level, we adopt a cross-attentive feature alignment strategy to learn the well-aligned fine-grained human features for adaptation. To better exploit the human topological structure, we model the human-topology structure via GCN and conduct cross-graph topology alignment across domains to preserve the structure-invariant knowledge. In SSDA setting, we additionally propose to adapt the corresponding keypoint heatmap representations to reduce the intra-domain gap. Extensive experiments show that our method significantly boosts performance of the target domain even with no labels or sparse labels. We also hope our method could inspire more ideas on the cross-domain pose estimation field in the future. 
\begin{acks}
This research is supported by the Fundamental Research Funds for the Central Universities (2020YJS037).
\end{acks}
{
\bibliographystyle{ACM-Reference-Format}
\bibliography{egbib}
}
\end{document}